\documentclass[11pt]{article}
\usepackage[utf8]{inputenc}
\usepackage[nohead, nomarginpar, margin=1.0in, foot=0.50in]{geometry}
\usepackage{amsmath,amssymb}
\usepackage{amsthm}
\usepackage{relsize}
\usepackage{mathtools}
\usepackage{bbm}
\usepackage{tikz}
\usepackage{wasysym}
\usepackage{comment}
\usepackage{array}
\usepackage[title]{appendix}
\newcolumntype{P}[1]{>{\centering\arraybackslash}p{#1}}
\usepackage[
backend=biber,
style=alphabetic,
sorting=ynt, maxbibnames=99
]{biblatex}
\newcommand\norm[1]{\left\lVert#1\right\rVert}
\newcommand*\circled[1]{\tikz[baseline=(char.base)]{
            \node[shape=circle,draw,inner sep=0.2pt] (char) {#1};}}
\newcommand{\defeq}{\vcentcolon=}

\DeclarePairedDelimiter\floor{\lfloor}{\rfloor}

\newtheorem{theorem}{Theorem}[section]

\newtheorem{lemma}[theorem]{Lemma}

\newtheorem{definition}[theorem]{Definition}
\newtheorem{remark}[theorem]{Remark}

\newtheorem{assumption}{Assumption}[section]

\DeclareMathOperator*{\argmin}{arg\,min}

\title{Universal Consistency of Wide and Deep ReLU Neural Networks and Minimax Optimal Convergence Rates for Kolmogorov-Donoho Optimal Function Classes}
\author{Hyunouk Ko and Xiaoming Huo}
\date{}

\begin{document}

\maketitle

\begin{abstract}
In this paper, we prove the universal consistency of wide and deep ReLU neural network classifiers trained on the logistic loss. 
We also give sufficient conditions for a class of probability measures for which classifiers based on neural networks achieve minimax optimal rates of convergence. 
The result applies to a wide range of known function classes. In particular, while most previous works impose explicit smoothness assumptions on the regression function, our framework encompasses more general settings. The proposed neural networks are either the minimizers of the logistic loss or the $0$-$1$ loss. 
In the former case, they are interpolating classifiers that exhibit a benign overfitting behavior. 
\end{abstract}

\section{Introduction}
\label{submission}
While the development of statistical theory for binary classification dates back to the 1970s and is well-summarized in \cite{devroye2013probabilistic} and \cite{boucheron2005theory}, a general theory explaining the generalizability of classifiers based on neural networks is far from complete. The problem can be roughly formulated as follows. The random vector $(X,Y)$ takes values in $\mathbb{R}^d\times \{0,1\}$, and we have $n$ independent, identically distributed samples $\{(X_1,Y_1),\dots,(X_n,Y_n) \}$. The goal is to build a function $g:\mathbb{R}^d \rightarrow \{0,1\}$ based on $n$ samples such that the classification risk of $g$, $E[g(X)\neq Y]$, is minimal. The function $\eta(x) = E[Y|X=x]$ is called the regression function. It is well-known that the Bayes classifier defined by 
\begin{align*}
    g^*(\boldsymbol{x}) \defeq
    \begin{cases}
        1 &\text{ if } \eta(x)\geq \frac{1}{2};\\
        0 &\text{ otherwise}
    \end{cases}
\end{align*}
achieves the minimal classification risk, $L^* \defeq E[g^*(X)\neq Y]$. Thus, it is natural to study the non-negative excess risk $E[g(X)\neq Y] - L^*$ of a classifier $g$ as a measure of its performance.

The first classical result on classifiers based on neural networks is the paper \cite{farago1993strong}, which establishes two results. First, it shows that there exists a sequence of 1-hidden layer sigmoidal neural network classifiers whose widths grow in the order $o\left(\frac{n}{\log n}\right)$ such that their excess risks converge to $0$ uniformly over all possible distributions, i.e., they are universally consistent. Second, for distributions whose regression function belongs to the Barron space \cite{barron1993universal}, a wide class of functions for which shallow neural networks enjoy dimension-free approximation rate, it is shown that there exist neural network classifiers whose excess risks converge at a uniform rate $O(n^{-\frac{1}{4}})$. 

However, the first result has room for improvement because it does not apply to deep or wide neural networks, and the proposed classifier is computationally infeasible. 
The second result on rates of convergence may be tightened in that there is no indication of whether the rate is tight in such a regime. 

In practice, state-of-the-art neural networks have become increasingly more complex with number of parameters employed scaling to the order of hundreds of trillions. 
A very recent work \cite{radhakrishnan2023wide} studied the weak consistency of infinitely wide and deep neural networks using polynomial and sinusoidal activation functions, interpreting them as the neural tangent kernel (NTK) machines. However, they leave the question of weak consistency, let alone strong consistency, of finitely wide and deep neural networks as an open problem (see Section 3 in their paper). \textbf{Our result answers this question and provides a theoretical guarantee that for an arbitrary distribution, a computationally feasible sequence of classifiers based on deep and wide neural networks is strongly consistent}.

A number of recent results study classification problems with overparametrized deep neural networks. \cite{kim2021fast} shows that in the classical regimes characterized by H\"{o}lder-smoothness, neural networks that minimize the empirical risk of the hinge loss or logistic loss achieve competitive rates of convergence. \cite{bos2022convergence} considers multi-class classification under the smooth compositional structural assumption on the regression function. Others derive convergence rates of convolutional neural networks optimizing the square loss \cite{kohler2022rate} and the logistic loss \cite{kohler2020statistical}. 

A common assumption employed above and in the classical statistics literature including \cite{mammen1999smooth}, \cite{tsybakov2004optimal}, \cite{audibert2007fast}, \cite{kerkyacharian2014optimal}, is to impose a smoothness assumption on the regression function (see Section 2.3 of \cite{suh2024survey} for details). Then, they derive upper bounds on the convergence rate and in some regimes, prove minimax optimality by deriving a matching minimax lower bound.

We take a somewhat different view and \textbf{ask in what distributional regimes neural network classifiers are capable of achieving minimax optimal rates}. In doing so, we relax the smoothness assumption on the regression function and allow for a study of much more general classes of $L^2$ functions. 

Specifically, we consider a family of $L^2$ functions with a finite Kolmogorov-Donoho optimal exponent, which is an information-theoretic number that quantifies the number of bits needed to construct an encoder-decoder pair that can approximate a given function class to a target accuracy (details in Section \ref{Subsection: kolmogorov-donoho}). \textbf{The significance of this characterization is that it applies to a much wider class of functions without explicit smoothness constraints, allowing for more realistic distributional settings}. A series of works from the past two decades (\cite{donoho1998data},\cite{grohs2023phase},\cite{hinrichs2008degree},\cite{petersen2018optimal}) have provided optimal exponents for many general classes of functions including $L^p$-Sobolev spaces, Besov spaces, bounded variation spaces, modulation spaces, and cartoon functions. Moreover, \cite{elbrachter2021deep} shows that most of these spaces are well-approximated by neural networks from the perspective of distortion theory (details in Section \ref{Subsection: kolmogorov-donoho}).

A notable characteristic of the proposed neural network classifiers is that they interpolate the data. We show that this is true in the case a surrogate loss is used for minimization. This feature is characteristic of neural networks trained in practice and is known as benign overfitting.

To put our work into context, we discuss some related works on benign overfitting of neural networks. 
\cite{pmlr-v178-frei22a} showed that two-layer neural networks with smoothed leaky ReLU activations trained with gradient descent exhibit exponentially fast convergence rates for distributions (roughly) with strongly log-concave covariate distribution and regression function whose norm is bounded by $1$. 
\cite{cao2022benign} also derived exponential rates for convolutional neural networks under assumptions that imply the regression function is binary-valued. 
\cite{kou2023benign} obtained similar results for a slightly more general setting with ReLU convolutional neural networks. 
We emphasize that the distributional assumptions in most related works are quite restrictive compared to our flexible distributional setting.

One important novelty in our proofs is how we control the estimation error. A typical way to do this has been to use the so-called calibration or comparison inequalities that bound the excess classification risk by some power of the excess surrogate risk. The first result of this type was Zhang's inequality \cite{zhang2004statistical}, which was later refined by \cite{bartlett2006convexity} under the relaxed condition of Tsybakov noise assumption (see also, Section 4.2, 5.2 of \cite{boucheron2005theory}, Chapters 3,8 of \cite{steinwart2008support}). 
However, as noted on page 15 of \cite{zhang2023classification}, this approach can yield suboptimal rates in certain regimes. 

Our approach overcomes this suboptimality by taking advantage of two observations: first, some neural networks can achieve exponentially fast rates of convergence for H\"older smooth functions, and second, $n$ i.i.d. random vectors are separated by a distance of at least $\Omega(n^{-\frac{2}{d}})$ with high probability, which allows for the construction of a H\"older smooth function with bounded norm that interpolates the signs of all $n$ points.

To summarize, we first show the universal consistency of wide and deep ReLU neural networks and second, give a characterization of some general classes of distributions for which neural network classifiers achieve minimax optimal rates of convergence.

\subsection{Organization}
In Section \ref{Section: preliminaries}, we give a rigorous formulation of binary classification problems, provide definitions involving neural networks, and introduce basic concepts from Kolmogorov-Donoho approximation theory.  
In Section \ref{Section: universal consistency}, we establish our first main result on the universal consistency of wide and deep ReLU neural networks. In Section \ref{Section: convergence rates}, we give our second main results on rates of convergence for neural network classifiers for functions with Kolmogorov-Donoho optimal exponents and demonstrate with examples how the theorems may be applied to specific function spaces.

\section{Preliminaries}\label{Section: preliminaries}

We first give a rigorous formulation of the classification problem. Suppose we have $Z=(X,Y)$ and $Z_i = (X_i,Y_i), i=1,2,\dots$ countably infinite, independent, identically distributed random vectors that map from a common probability space $(\Omega, \Sigma, P)$ to $[0,1]^d \times \{0, 1\}$. 

Fix a positive integer $n$. By a \textbf{classifier}, we mean a measurable function $g_n: [0,1]^d \times \{[0,1]^d \times \{0,1\}\}^n \rightarrow \{0, 1\}$ where $[0,1]^d$ is endowed with the usual Borel $\sigma$-algebra it inherits from $\mathbb{R}^d$. Then, we can define
\begin{align*}
     L(g_n) \defeq P(g_n(X,Z_1,\dots,Z_n)\neq Y | Z_1,\dots, Z_n)
\end{align*}
which is the conditional probability with respect to the $\sigma$-algebra generated by $Z_1,\dots, Z_n$. Note that $L(g_n)$ is well-defined up to $P$-null set and is  $\sigma(Z_1,\dots,Z_n)$-measurable by the Radon-Nikodym theorem. For $n=0$, we let $L_0=L(g)=P(g(X)\neq Y)$ in the obvious way. We will be interested in $E[L(g_n)]$, the classification risk, as a measure of the performance of a classifier $g_n$.

Given a real-valued function $f:  [0,1]^d \times \{[0,1]^d \times \{0,1\}\}^n  \rightarrow \mathbb{R}$, the \textbf{plug-in classifier} corresponding to $f$ will be defined as:
\begin{align} \label{eq6}
    p_f(\boldsymbol{x}) \defeq \mathbbm{1}_{\{x:f(x)\geq 1/2\}}(\boldsymbol{x}),
\end{align}
where for any subset $A\subset\mathbb{R}^d$, $\mathbbm{1}_{A}$ is the indicator function defined by
\begin{align}
    \mathbbm{1}_{A}(x) \defeq
    \begin{cases}
        1, & \text{ if } x\in A;\\
        0, & \text{ otherwise.}
    \end{cases}
\end{align}
When clear from context, we will also write $L(f) \defeq L(p_f)$.

Denote by $\mathbb{N}$ the set of natural numbers $\{1,2,\dots\}$. Any sequence of valid classifiers $\{g_n\}_{n \in \mathbb{N}}$ will be called a \textbf{classification rule}. A classification rule will be called weakly consistent if $L(g_n) \rightarrow L^*$ in probability (equivalently, $E[L(g_n)] \rightarrow L^*$) and strongly consistent if $L(g_n) \rightarrow L^*$ almost surely. Note these notions depend on the underlying probability measure $P$. 
We will call a classification rule universally weakly (strongly) consistent if for all valid probability measures $P$, the rule is weakly (strongly) consistent.

\subsection{Notations}\label{Subsection: notations}
The symbols $\mathbb{Z}, \mathbb{R}$ denote the set of integers and real numbers respectively, and $\mathbb{R}_{>0}$ denotes the positive real numbers. For any $x\in \mathbb{R}$, we define $\floor{x} \defeq \max\{m \in \mathbb{Z}: m\leq x\}$.
We write $L^p([0,1]^d,\mu)$ or $L^p(\mu)$ to denote the $L^p$ space with respect to a positive Borel measure $\mu$. 
This metric space has the usual $L^p(\mu)$-norm and has the corresponding metric topology. We write $C([0,1]^d)$ to denote the space of all continuous functions on $[0,1]^d$ equipped with the uniform norm, $\norm{f}_u \defeq \sup_{x\in[0,1]}|f(x)|$, and the usual norm topology. For an integer $k\geq0$ and $0<\beta\leq 1$, we define the H\"older space $C^{k,\beta} = C^{k,\beta}([0,1]^d)$ as the space of all $k$-times continuously differentiable functions on $[0,1]^d$ equipped with the norm:
\begin{align*}
    \norm{f}_{C^{k,\beta}} = &\max \biggl\{ \max_{\mathbf{k}: |\mathbf{k}|\leq k} \max_{\mathbf{x} \in [0,1]^d} |D^{\mathbf{k}}f(\mathbf{x})|,\\ 
    &\max_{\mathbf{k}: |\mathbf{k}|=k} \sup_{\substack{\mathbf{x},\mathbf{y} \in[0,1]^d\\ \mathbf{x}\neq \mathbf{y}}} \frac{\norm{D^{\mathbf{k}}f(\mathbf{x}) - D^{\mathbf{k}}f(\mathbf{y})}_2}{\norm{\mathbf{x}-\mathbf{y}}_2^{\beta}} \biggl\}.
\end{align*}
For either a matrix $A\in\mathbb{R}^{m\times n}$ or a vector $v\in\mathbb{R}^n$, $\norm{A}_{\text{max}}\defeq \max_{i=1,\dots,m}\max_{j=1,\dots,n}|A_{ij}|$ and $\norm{v}_{\text{max}} \defeq \max_{i=1,\dots,n}|v_i|$ where the subscript notation refers to the indexed component of the matrix and vector. For a real-valued measurable function $f$ whose domain is a measurable space $(\Omega, \Sigma, P)$, we write $P(f)$ to denote the integral of $f$ with respect to $P$. For a probability measure $P$, we will write $P_n$ to mean the empirical measure corresponding to $n$ i.i.d. random variables with distribution $P$, $\frac{1}{n} \sum_{i=1}^n \delta_{X_i}(B)$ where $\delta_{X}(B) = \mathbbm{1}_{B}(X)$ for any measurable set $B$. For a metric space $S$, $\mathcal{B}(S)$ denotes the Borel $\sigma$-algebra associated with $S$.

\subsection{Neural networks}\label{Section: neural networks}

In this section, we rigorously define neural networks and their realization functions and equip the space with the right topology to obtain an adequate compactification of the space of neural networks.

Fix $L, N_0,\dots, N_L \in \mathbb{N}$. We define a \textbf{neural network} as the ordered set of matrix-vector tuples $\Phi = \{(A_l,b_l)\}_{l=1}^L$ where $A_l \in \mathbb{R}^{N_l\times N_{l-1}}$ and $b_l \in \mathbb{R}^{N_l}$. We call the ordered tuple $S=(L,N_0,\dots,N_L)$ the \textbf{architecture} of $\Phi$. We define $\mathcal{NN}(S)$ to be the set of all neural networks with architecture $S$. We sometimes write  $\mathcal{NN}_{d,1}(S)$ to make explicit the restriction that the $N_0=d, N_L=1$. That is two neural networks $\Phi_1, \Phi_2$ belong to the same $\mathcal{NN}(S)$ if and only if the dimensions of all the matrices and vectors defining them match. When a neural network $\Phi$ is given, we write $S(\Phi)$ to denote its architecture. In the rest of the paper, we will only be concerned with the case $N_0 =d, N_L=1$.

Now let $\varrho:\mathbb{R} \rightarrow \mathbb{R}$ be the ReLU activation function $\varrho(x) \defeq \max \{x,0\}$. For a vector $v=(v_1,\dots,v_n) \in \mathbb{R}^n$, with a slight abuse of notation, we write $\varrho(v)$ to mean $\varrho(v) \defeq (\varrho(v_1), \dots, \varrho(v_n))\in \mathbb{R}^n$. Also, let $\mathcal{NN} \defeq \bigcup_{S}\mathcal{NN}(S)$ where the union runs over all choices of valid architectures $S$. For a given set $\Omega \subset \mathbb{R}^{N_0}$, we can define the realization map of a neural network $\Phi$ as the map $R_{\varrho}^{\Omega}: \mathcal{NN} \rightarrow C(\Omega)$ where $R_{\varrho}(\Phi): \Omega \rightarrow \mathbb{R}$ is defined in the following recursive fashion:
\begin{align*}
    &R_{\varrho}(\Phi)(x) = x_L \text{ where}\\
    &x_0 \defeq x\\
    & x_l \defeq \varrho(A_l x_{l-1} + b_l), l=1,\dots,L-1\\
    & x_L \defeq A_Lx_{L-1} + b_L.
\end{align*}
For a given architecture $S$, We will define the total number of neurons as $N(S)\defeq \sum_{i=1}^L N_i$, and the number of layers as $L(S) \defeq |S|$ where $|S|$ is the cardinality of $S$. 
Furthermore for a given $\Phi \in \mathcal{NN}(S)$, we define the following quantities that specify the complexity of $\Phi$:
\begin{itemize}
    \item the connectivity $\mathcal{M}(\Phi)$ denotes the total number of nonzero entries in the matrices $A_{\ell}, \ell \in\{1,2, \ldots, L\}$, and the vectors $b_{\ell}, \ell \in\{1,2, \ldots, L\}$,
    \item width $\mathcal{W}(\Phi):=\max _{\ell: 0\leq \ell \leq L} N_{\ell}$,
    \item $\mathcal{L}(\Phi)$ is the total number of hidden layers in the architecture defining $\Phi$,
    \item weight magnitude \\$\mathcal{B}(\Phi):=\max _{\ell: 0\leq \ell \leq L} \max \left\{\left\|A_{\ell}\right\|_{\text{max}},\left\|b_{\ell}\right\|_{\text{max}}\right\}$.
\end{itemize}

We also make $\mathcal{NN}(S)$ a finite-dimensional normed space by equipping it with the norm 
$$\norm{\Phi}_{\mathcal{NN}} \defeq \max_{\ell: 0\leq \ell \leq L} \norm{A_l}_{\text{max}} + \max_{\ell: 0\leq \ell \leq L} \norm{b_l}_{\text{max}}.$$

For a fixed architecture $S$ and a fixed choice of function $\pi:\mathbb{N} \rightarrow \mathbb{R}_{>0}$, we define $\mathcal{NN}^{\pi,S}_{d,1}(M)$ to be the class of neural networks with architecture $S$ and whose weights are bounded by $\pi(M)$:
\begin{align}\label{neural network class 1}
    \mathcal{NN}^{\pi,S}_{d,1}(M) \defeq \{\Phi \in \mathcal{NN}_{d,1}(S): \mathcal{M}(\Phi) \leq M,\\
    \mathcal{B}(\Phi) \leq \pi(M) \}.
\end{align}
for any $M>0$. 

It is possible to give a partial-order $\leq$ to the set of architectures by stipulating $S_1 \leq S_2$ for $S_1=(L,N_1,\dots, N_L)$ and $S_2=(L',M_1,\dots, M_{L'})$ if and only if $L\leq L'$ and $N_i \leq M_i$ for all $i=1,\cdots,L$.

For the purposes of proving universal consistency in Section \ref{Section: universal consistency}, we want to consider a method of sieves where we choose an estimator $\widehat{\theta}_n$ from $\mathcal{NN}^{\pi,S_n}_{d,1}(M_n)$ for a suitable choice of increasing sequence of architectures $\{S_n \}_{n\in\mathbb{N}}$ and real numbers $\{M_n\}_{n\in\mathbb{N}}$.  Therefore, our neural networks will come from the set of a countable union:
\begin{align*}
    \bigcup_{n=1}^{\infty} \mathcal{NN}^{\pi,S_n}_{d,1}(M_n).
\end{align*}
We want to give this set a topology so that we have a compact space: this is necessary to apply Wald's method for proving consistency. Thus, we consider the following construction in the next two paragraphs.

For each $n\in\mathbb{N}$, let $d_{n}(\cdot, \cdot)$ be the metric on $\mathcal{NN}^{\pi,S_n}_{d,1}(M_n)$ induced by the norm $\norm{\cdot}_{\mathcal{NN}}$.  Then, define the disjoint union space: 
\begin{align}
    \widetilde{\Theta} \defeq \bigsqcup_{n=1}^{\infty} \mathcal{NN}^{\pi,S_n}_{d,1}(M_n)
\end{align}
with the disjoint union topology. This space is also metrizable and so normal. 
We can give an explicit metric that metrizes this topology: if we let $D_n$ be the diameter of the space $\mathcal{NN}^{\pi,S_n}_{d,1}(M_n)$ for all $n\in\mathbb{N}$,
\begin{align}\label{metric}
    d(x,y) = 
    \begin{cases}
        d_n(x,y), &  \text{if } x,y \in \mathcal{NN}^{\pi,S_n}_{d,1}(M_n);\\
        \max \{D_n,D_m \}, &  \text{if } x\in \mathcal{NN}^{\pi,S_n}_{d,1}(M_n);\\
        &y \in \mathcal{NN}^{\pi,S_m}_{d,1}(M_m), n\neq m
    \end{cases}
\end{align}
is such a metric (c.f. Example 2.6, Theorem 2.12 of \cite{sharma2020disjoint}). 
It is a second-countable, complete metric space. Since it is the disjoint union of countably many compact Hausdorff spaces, it is also a locally compact Hausdorff space. 

The above construction ensures the existence of the Stone-\v{C}ech compactification of $\widetilde{\Theta}$, which we denote by $\Theta$. Recall that the Stone-\v{C}ech compactification is characterized by the fact that $\Theta$ is a compact Hausdorff space containing $\widetilde{\Theta}$ as a dense subspace and that any continuous function $f: \widetilde{\Theta} \rightarrow C$ for any compact Hausdorff space $C$ can be uniquely extended to a continuous function $\Bar{f}: \Theta \rightarrow C$. This compactification is unique up to equivalence that identifies two compactifications $Y_1$, $Y_2$ of $\widetilde{\Theta}$ such that there exists a homeomorphism $h: Y_1 \rightarrow Y_2$ that is an identity when restricted to $\widetilde{\Theta}$. In fact, $\Theta$ is not metrizable because $\widetilde{\Theta}$ is non-compact. One point of caution is that while all points of $\Theta \backslash \widetilde{\Theta} $ are limit points of $\widetilde{\Theta}$ by definition of compactification, none of them are a (sequential) limit of any sequence of points from $\widetilde{\Theta}$.

It is not difficult to check that the realization mapping $R_{\varrho}: \widetilde{\Theta} \rightarrow C(\Omega)$ is continuous when $C(\Omega)$ is equipped with the uniform norm (for e.g., Proposition 4.1 of \cite{petersen2021topological}). For our analysis, we may assume without loss of generality that the realization mapping is followed by a projection to the unit ball in $C(\Omega)$, which we denote by $U(C(\Omega))$. This map is also continuous because the projection is achieved by mere scaling. Furthermore, we extend the domain of the realization mapping to $\Theta$, which is possible by the characterizing property of Stone-Cech compactification. 

We will also need to consider a more general class for the results in Section \ref{Section: convergence rates}. We define $\widetilde{\mathcal{NN}}_{d,1}$ to be a directed acyclic graph with input-dimension $d$ and output dimension $1$ such that all nodes can be grouped into $L$ layers and connections of a node at a given layer may come from any of the earlier layers but not from the same layer. 
Other definitions involving a neural network such as width, depth, and connectivity, still remain valid.
Moreover, we will allow the non-linear activations used in the realization function to be either the ReLU $\varrho$ or a Lipschitz continuous, periodic function $\widetilde{\varrho}$ with period $T>0$ that satisfies:
\begin{align*}
    \widetilde{\varrho}(x) &>0 \text{ for all } x\in(0,T/2),\\
    \widetilde{\varrho}(x) &<0 \text{ for all } x\in(T/2,T),\\
    \max_{x\in\mathbb{R}}\widetilde{\varrho}(x) &= -\min_{x\in\mathbb{R}}\widetilde{\varrho}(x).
\end{align*}
Then, we write $R_{\varrho,\widetilde{\varrho}}(\Phi)$ for a realization of a network $\Phi\in \widetilde{NN}_{d,1}$ with mixed choice of activation functions allowed.

\subsection{Kolmogorov-Donoho approximation theory}\label{Subsection: kolmogorov-donoho}

In this subsection, we introduce the concepts from the Kolomogorov-Donoho approximation theory that appear in Section \ref{Section: convergence rates}. In particular, we assume that the regression function belongs to a function class with an information-theoretic constraint.

Let $l \in \mathbb{N}$, $d \in \mathbb{N}$, $\Omega \subset \mathbb{R}^d$ such that $\Omega$ is Lebesgue measurable. In all that follows, we equip $\Omega$ with the Borel $\sigma$-algebra and the $d$-dimensional Lebesgue measure on it. Let $\mathcal{C}$ be a class of functions $\mathcal{C} \subset L^2(\Omega)$.  First, define the set of encoders and the set of decoders as follows:
\begin{align*}
    \mathcal{E}^{l} \defeq \{ E: \mathcal{C} \rightarrow \{0,1\}^l\},\\
    \mathcal{D}^{l} \defeq \{ D: \{0,1\}^l \rightarrow \mathcal{C} \}.
\end{align*}

\begin{definition}[Kolmogorov-Donoho optimal exponent]\label{optimal exponent}
    For each $\epsilon>0$, let the minimax code length be defined as:
    \begin{align*}
        L(\epsilon, \mathcal{C}):=  \min \{ &\ell \in \mathbb{N}: \exists(E, D) \in \mathfrak{E}^{\ell} \times \mathfrak{D}^{\ell}:. \\
        & \sup_{f \in \mathcal{C}}\|D(E(f))-f\|_{L^2(\Omega)} \leq \epsilon \} .
    \end{align*}
    We define the (Kolmogorov-Donoho) optimal exponent of $\mathcal{C}$ as the real number
    \begin{align}
        &\gamma^*(\mathcal{C}) \defeq \sup \{\gamma \in \mathbb{R}: L(\epsilon, \mathcal{C}) \in O(\epsilon^{-\frac{1}{\gamma}}) \}. \nonumber
    \end{align}
\end{definition}
The optimal exponent is known for $L^p$-Sobolev spaces, Besov spaces, modulation spaces, and Cartoon function classes as summarized in Table 1 of \cite{elbrachter2021deep}.

There is a rich literature on the class of basis functions whose linear combinations can be used as approximators for these function spaces. That is, given a Hilbert space $\mathcal{H} = L^2(\Omega)$ for some bounded set $\Omega \subset \mathbb{R}^d$, we consider a countable family of functions in $\mathcal{H}$, called a dictionary and denoted $\mathcal{D}=\{\psi_{i} \}_{i\in \mathbb{N}}$, with which we approximate any function from $\mathcal{C} \subset \mathcal{H}$. We may measure the performance of $\mathcal{D}$ with respect to $\mathcal{C}$ with the following quantity:
\begin{align}
    \varepsilon^{\pi}_{\mathcal{C},\mathcal{D}}(M) \defeq \sup_{f \in \mathcal{C}} \inf_{\substack{I_{f,M} \subset \{1,2,\dots, \pi(M)\}\\ |I_{f,M}|=M, |c_i|\leq \pi(M)}} \norm{f-\sum_{i\in I_{f,M}}c_i\psi_i}_{L^2(\Omega)}
\end{align}
where $\pi$ denotes some given real polynomial.  
Then, one defines the effective best $M$-term approximation rate of $\mathcal{C}$ with dictionary $\mathcal{D}$ as: 
\begin{definition}[Effective best $M$-term approximation rate]
    \begin{align*}
        \gamma^{*}(\mathcal{C},\mathcal{D}) \defeq \sup \{&\gamma\geq 0: \exists \text{ polynomial } \pi \text{ such that } \\
        &\varepsilon^{\pi}_{\mathcal{C},\mathcal{D}}(M) \in O(M^{-\gamma}) \}.
    \end{align*}
\end{definition}
A notable relationship between $\gamma^*(\mathcal{C})$ and $\gamma^*(\mathcal{C}, \mathcal{D})$ is $\gamma^*(\mathcal{C}, \mathcal{D}) \leq \gamma^*(\mathcal{C})$. Then, we say that \textit{$\mathcal{C}$ is optimally representable by $\mathcal{D}$ if $\gamma^*(\mathcal{C}, \mathcal{D}) = \gamma^*(\mathcal{C})$}. 
Many function spaces usually studied in the approximation theory literature are, in fact, optimally representable by well-known dictionaries such as those based on the Fourier/wavelet basis and the Haar basis. 

There is a natural corresponding concept for the class of neural networks as a replacement for dictionaries. Recalling the definition \eqref{neural network class 1}, we will define the union of all neural networks whose architecture has depth bounded by $\pi(M)$ for a given function $\pi$. Specifically,
\begin{align*}
    \mathcal{NN}^{\pi}_{d,1}(M) \defeq \bigcup_{S: L(S)\leq \pi(\log M)}\mathcal{NN}^{\pi, S}_{d,1  }(M).
\end{align*}
Note for this definition, we don't care about the topology on this set at this point.

Similar to the effective best approximation error $\varepsilon^{\pi}_{\mathcal{C},\mathcal{D}}(M)$, defined with respect to the dictionary $\mathcal{D}$, we define the effective best approximation with neural networks as follows:
\begin{align}
    \varepsilon^{\pi}_{\mathcal{N}}(M) \defeq \sup_{f \in \mathcal{C}} \inf_{\Phi \in \mathcal{NN}^{\pi}_{d,1}(M)} \norm{f-R_{\varrho}(\Phi)}_{L^2(\Omega)}.
\end{align}
Just as we did for the dictionary $\mathcal{D}$, we define the best effective $M$-term approximation rate as follows:
\begin{definition}[Effective best $M$-weight approximation rate]
    \begin{align*}
        \gamma^{*}_{\mathcal{N}}(\mathcal{C}) \defeq \sup \{&\gamma\geq 0: \exists \text{ polynomial } \pi \text{ such that }\\
        &\varepsilon^{\pi}_{\mathcal{N}}(M) \in O(M^{-\gamma}), M\rightarrow \infty \}. 
    \end{align*}
\end{definition}
This means if $\gamma^{*}_{\mathcal{N}}(\mathcal{C})>0$, the $L^2$ approximation error decays at least polynomially in the connectivity of the approximating neural networks. Furthermore, it is shown in Theorem VI.4 of \cite{elbrachter2021deep} that $\gamma^{*}_{\mathcal{N}}(\mathcal{C}) \leq \gamma^*(\mathcal{C})$, which makes the following definition natural:
\begin{definition}
    We say that $\mathcal{C} \subset L^2(\Omega)$ is optimally representable by neural networks if 
    \begin{align*}
        \gamma^{*}_{\mathcal{N}}(\mathcal{C}) = \gamma^*(\mathcal{C}).
    \end{align*}
\end{definition}
Quite general classes of functions are optimally representable by neural networks including the Besov spaces and the modulation spaces. These results follow from the ``transference principle" which shows that $\gamma^{*}(\mathcal{C},\mathcal{D}) \leq \gamma^{*}_{\mathcal{N}}(\mathcal{C})$ for most useful dictionaries that optimally represent classical function spaces.

\section{Universal consistency}\label{Section: universal consistency}

In this section, we state our first result on the universal consistency of wide and deep ReLU neural network classifiers.

We will need the following lemma to establish that the empirical risk minimizer is well-defined as a classifier. Its proof is relegated to Appendix \ref{proof1}.

\begin{lemma}\label{measurability lemma}
    Let $(A,\mathcal{A})$ be a measurable space and $B$ a compact, metrizable topological space. Assume $m(\cdot,\cdot): A \times B \rightarrow \mathbb{R}$ is measurable in the first argument and continuous in the second argument. Then, there exists a Borel measurable mapping  $\widehat{f}: A \rightarrow B$ that satisfies $m(a,\widehat{f}(a)) = \sup_{b \in B} m(a,b)$ is Borel measurable. 
\end{lemma}

Now, we state our main theorem on the universal consistency of wide and deep ReLU neural networks. Its proof is relegated to Appendix \ref{proof2}.

\begin{theorem}\label{Theorem: universal consistency}
    Let $\{S_n\}_{n\in \mathbb{N}}$ be an increasing sequence of architectures such that $W(S_n)\geq n$ or $L(S_n)\geq n$ for all $n\in\mathbb{N}$. There exists some increasing function $\pi$ and constant $c_d$ only depending on $d$ such that the empirical risk minimizer of the logistic loss on $\mathcal{NN}^{\pi,S_n}_{d,1}(c_d n)$ for each $n\in \mathbb{N}$ defined as:
    \begin{align}\label{theorem1:eq1}
        \widehat{\theta}_n \defeq \argmin_{\theta \in \mathcal{NN}^{\pi,S_n}_{d,1}(n)} \frac{1}{n}\sum_{i=1}^n l(R_{\varrho}(\theta)(X_i),Y_i)
    \end{align}
    is universally strongly consistent:
    \begin{align*}
        \lim_{n\rightarrow \infty} L(R_{\varrho}(\widehat{\theta}_n)+1/2) \rightarrow L^* \text{ with probability } 1.
    \end{align*}
\end{theorem}

\begin{remark}
    As can be seen from the proof, the only property that we require of the surrogate loss is that its empirical minimizer in $\mathcal{NN}_{d,1}^{\pi,S_n}(c_d n)$ achieves $0$ classification loss. The same conclusion holds for any other continuous loss function with such property.
\end{remark}

As noted in the Introduction, this result answers the open problem mentioned in \cite{radhakrishnan2023wide}. In fact, the classifiers in Theorem \ref{Theorem: universal consistency} are interpolating classifiers, i.e., they correctly classify all training points, and are also feasible as they are the minimizers of a convex surrogate loss. 

\section{Rates of convergence}\label{Section: convergence rates}
The second question of interest, which is more practically relevant, is what upper bounds we can establish on the excess risk of the empirical risk minimizer \eqref{theorem1:eq1} as a function of $n$ that is independent of any individual choice of the underlying distribution, i.e., we want to establish a uniform (in the set of probability measures) rate of convergence.  It is well known that no universal rates that hold for all probability distributions are possible (c.f. Theorem 7.2 of \cite{devroye2013probabilistic}). 

This means that we must have some restrictions on the set of possible $P$. Observing that the joint distribution of $(X,Y)$ on $[0,1]^d \times \{0,1\}$ is fully determined by the specification of $E[Y|X]$ and the marginal measure $\mu_X$ on $[0,1]^d$, we take the view of considering all $P$ such that the regression function belongs to some given model class of functions and the marginal law of $X$ satisfies certain regularity conditions.

What model classes are suitable and interesting for practical relevance is in itself an important question. As noted in the Introduction, smoothness assumptions are most widely used. We generalize the landscape of classification theory by taking advantage of how well neural networks can approximate the most useful dictionaries. 

Our program will work with the usual decomposition of the excess risk in terms of estimation and approximation error:

\begin{align*}
    \mathcal{E}(\widehat{f}_n) = \underbrace{E[L_n] - \inf_{f \in \mathcal{F}_n}E[L(f)]}_{\circled{1}} + \underbrace{\inf_{f \in \mathcal{F}_n}E[L(f)] - L^*}_{\circled{2}}
\end{align*}

where term $\circled{1}$ comprises the estimation error, and we will rely on empirical risk minimization and more fundamentally, empirical process theory to control this error. Term $\circled{2}$ comprises the approximation error, and we control it by proposing suitable classes of neural networks that well-approximate the regression function n $L^p$ (c.f. Section \ref{regression and classification}).

\subsection{Distributional assumptions}\label{Section: assumptions}
For our results on uniform convergence rates, we will make the following three assumptions:
\begin{assumption}(Tsybakov noise condition) \label{Assumption: 1}
    We assume there exist constants $C_0>0$ and $\alpha \geq 0$ such that
    \begin{align}\label{noise condition}
        P_X(0<|\eta(x)-1 / 2| \leq t) \leq C_0 t^\alpha, \quad \forall t>0. 
    \end{align} 
\end{assumption}

\begin{remark}
    This assumption is used widely in the literature and controls the concentration of measure near the optimal decision boundary. The assumption becomes vacuous for $\alpha=0$ and the case $\alpha=\infty$ corresponds to a strict margin condition.
\end{remark}

\begin{assumption}\label{Assumption: 2}
    We assume that the distribution of $X$ admits an $L^2$ density with respect to the $n$-dimensional Lebesgue measure restricted to $[0,1]^d$ that is uniformly bounded by some constant.
\end{assumption}

\begin{remark}
    While we have adopted the Lebesgue measure as the dominating measure of $P$ to take advantage of the known approximation results, we believe the approximation theory can be generalized to arbitrary $\sigma$-finite measures.
\end{remark}

\begin{assumption}\label{Assumption: 3}
    We assume that the regression function belongs to some class of functions $\mathcal{F} \subset L^2([0,1]^d)$ with a finite Kolmogorov-Donoho optimal exponent $\gamma^*(\mathcal{F})>0$.
\end{assumption}

\subsection{Convergence rates}
In this section, we give our second main results that characterize sufficient conditions for a set of probability measures under which neural network classifiers achieve minimax optimality.

There is a somewhat subtle relationship between regression and classification, and we relegate a detailed discussion on this relationship to Appendix \ref{regression and classification}.  
For now, we remark that while $L^p$ consistency is a sufficient but not a necessary condition for the consistency of the corresponding plug-in classification rule (pointwise regime), the convergence rate for the $p$th power of $L^p$ norm in the minimax sense for some classical function spaces may agree with the minimax rate of convergence for the classification risk.

We also remark that the observation of \cite{audibert2007fast} in the paragraph after Lemma 5.2 is somewhat misleading: the paper claims that deriving convergence rates for classification risk based on $L^2$ risk is not the right tool in the presence of Tsybakov noise condition. Specifically, under a suitable regime, for some constant $c>0$,
\begin{align*}
    \liminf _{n \rightarrow \infty} \inf _{T_n} \sup _{f \in \Sigma(\beta, L)} \mathbf{E}_f\left[n^{\frac{2 \beta}{2 \beta+d}}\left\|T_n-f\right\|_2^2\right] \geq c,
\end{align*}
where the infimum is over all possible estimators and $\Sigma(\beta,L)$ is the $L$-H\"{o}lder ball of functions, which then implies that inequality \eqref{classification and Lp norm} (in Appendix \ref{regression and classification}) only leads to suboptimal rates for the classification risk. However, while $n^{-\frac{2\beta}{2\beta+d}}$ is certainly the best possible rate for the square of $L^2$ risk in the above sense, it is only so when the infimum is taken over an estimator sequence ($T_n$'s), not deterministic functions. 

The approach using estimation and approximation error decomposition, on the other hand, allows us to fully use the approximation power of realizations of neural networks that lead to minimax optimal rates even in the presence of the Tsybakov noise condition.

Now, we state our results on the convergence rates of neural network classifiers in the framework of function classes with finite Kolmogorov-Donoho optimal exponents. For our first result, the empirical risk minimization is taken for the classification loss. Its proof is relegated to Appendix \ref{proof3}.

\begin{theorem}\label{Theorem: rate of convergence 1}
    Let $\mathcal{F}$ be a compact subset of $L^2([0,1]^d)$ with Kolomogorov-Donoho optimal exponent $\gamma^*>0$ that is optimally representable by neural networks with polynomial $\pi$.  Let $\mathcal{P}_{\mathcal{F}}$ be a given class of distributions satisfying Assumption \ref{Assumption: 1}, Assumption \ref{Assumption: 2}, and Assumption \ref{Assumption: 3} (with above $\mathcal{F}$). Define the optimal minimax rate of convergence for $\mathcal{P}_{\mathcal{F}}$ as follows:
    \begin{align*}
        m^* \defeq \inf\biggl\{m\in \mathbb{R}_{+}:& m \text{ satisfies } \inf_{g_n} \sup_{P\in \mathcal{P}_{\mathcal{F}}} E[L(g_n)] - L^*\\
        &= \Omega(n^{-m}) \biggr\}.
    \end{align*}
    Additionally, assume that $\alpha,\gamma^*,m^*$ satisfy 
    \begin{align} \
        2(1+\alpha)\gamma^*(1-m^*)&\geq (2+\alpha)m^*. \label{Theorem5.5: condition}
    \end{align}
    Define 
    \begin{align*}
        \mathcal{NN}_{n} \defeq \mathcal{NN}^{\pi}_{d,1}(C_{d,\alpha,m^*,\gamma^*}n^{\frac{(2+\alpha)m^*}{2(1+\alpha)\gamma^*}}).
    \end{align*}
    where $C_{d,\alpha,m^*,\gamma^*}$ is a constant that only depends on $d,\alpha,m^*,\gamma^*$ (see Definition \ref{optimal exponent}).
    Let $\widehat{\theta}_n$ be the empirical risk minimizer of the classification loss:
    \begin{align*}
        \widehat{\theta}_{n}  \defeq \argmin_{\theta \in \mathcal{NN}_n} \frac{1}{n} \sum_{i=1}^n P(p_{R_{\varrho}(\theta)}(X_i) \neq Y_i)
    \end{align*}
    Then the plug-in classification rule based on $\{R_{\varrho}(\widehat{\theta}_n)\}_{n\in\mathbb{N}}$ achieves minimax optimal (up to polylogarithmic factor) rate of convergence for the excess classification risk. 
\end{theorem}
\begin{remark}
    Note $m^*$ may depend on $\alpha$ and $\mathcal{F}$. Condition \eqref{Theorem5.5: condition} is not very stringent for many classical function spaces: Examples of classical regimes in which \eqref{Theorem5.5: condition} holds will be provided in Section \ref{Section: examples}. In fact, the condition turns out to be vacuous for the space of Besov functions.
\end{remark}

\begin{remark}
    Suppose $\mathcal{P}_{\mathcal{F}}$ is such that the minimax rate of $L^2$-risk matches that of classification risk in the sense of \eqref{yang minimax} and the rate is given by $n^{-m^*}$.
    Theorem \ref{Theorem: rate of convergence 1} shows that this optimal rate is still achieved if the infimum on the right-hand side of \eqref{yang minimax} is taken over all $f_n \in \mathcal{F}_n$ instead. Does this mean that we also get the same rate if we replace the left-hand side of \eqref{yang minimax} by $f_n\in\mathcal{F}_n$? Because $\mathcal{F}$ is optimally representable by neural networks with exponent $\gamma^*$, for any constant $C>0$ and any $m>\frac{(2+\alpha)m^*}{2(1+\alpha)}$ (in particular, $m=m^*$),
    \begin{align*}
        Cn^{-m} < \sup_{P \in \mathcal{P}} \inf_{f \in \mathcal{F}_n} (E_n[\norm{f-\eta}_2^2])^{\frac{1}{2}}
    \end{align*}
    happens infinitely often as $n \rightarrow \infty$. Because we have
    \begin{align*}
        \sup_{P \in \mathcal{P}_{\mathcal{F}}} \inf_{f \in \mathcal{F}_n} &(E_n[\norm{f-\eta}_2^2])^{\frac{1}{2}}\\
        &\leq \inf_{f_n\in\mathcal{F}_n}\sup_{P \in \mathcal{P}_{\mathcal{F}}} (E_n[\norm{f_n-\eta}_2^2])^{\frac{1}{2}},
    \end{align*}
    we conclude that the answer is no.
\end{remark}

The next result is analogous to Theorem \ref{Theorem: rate of convergence 1}, but now the empirical risk minimization is taken for the logistic loss over a class of feed-forward neural networks where skip-connections are allowed and two choices of activation functions (ReLU or a fixed piecewise linear, periodic function) are allowed (see end of Section \ref{Section: neural networks}). Under a slightly stronger condition, we show that the empirical risk minimizers of the logistic loss also achieve minimax optimality.

We emphasize that an important novelty in the proof is that the proposed neural networks, despite being trained on the logistic loss, achieve $0$ classification error on the training data. 
This is possible from the observation that $n$ i.i.d. points are separated as $\Omega(n^{-\frac{2}{d}})$ with high probability, from which we can show the existence of a H\"older smooth function that has correct signs on all $n$ points. Then, the approximation power of proper neural networks can be put to use. We now state our final result whose proof can be found in Appendix \ref{proof4}.

\begin{theorem}\label{Theorem: rate of convergence 2}
    Let $\mathcal{F}$ be a compact subset of $L^2([0,1]^d)$ with Kolomogorov-Donoho optimal exponent $\gamma^*>0$ that is optimally representable by neural networks with polynomial $\pi$.  Let $\mathcal{P}_{\mathcal{F}}$ be a given class of distributions satisfying Assumption \ref{Assumption: 1}, Assumption \ref{Assumption: 2}, and Assumption \ref{Assumption: 3} (with the above $\mathcal{F}$). Define the optimal minimax rate of convergence for $\mathcal{P}_{\mathcal{F}}$ as follows:
    \begin{align*}
        m^* \defeq \inf\biggl\{m\in \mathbb{R}_{+}: & m \text{ satisfies } \inf_{g_n} \sup_{P\in \mathcal{P}_{\mathcal{F}}} E[L(g_n)] - L^*\\ &\geq C_d n^{-m} \text{ for some constant } C_d \biggr\}.
    \end{align*}
    Additionally, assume that $\alpha,\gamma^*,m^*$ satisfy 
    \begin{align} 
        (1+\alpha)\gamma^*(1-m^*)&\geq (2+\alpha)m^*. \label{Theorem 5.7: condition 1}
    \end{align}
    Define 
    \begin{align*}
        \mathcal{NN}_{n} \defeq \biggl\{&\Phi \in \widetilde{\mathcal{NN}}_{d,1}(S):  \\
        &L(S), \mathcal{M}(\Phi) \leq C_{d,\alpha, \gamma^*} n^{\frac{(2+\alpha)m^*}{2(1+\alpha)\gamma^*}} \log(n+1) \biggr\}.
    \end{align*}
    where $C_{d,\alpha,m^*,\gamma^*}$ is a constant that only depends on $d,\alpha,m^*,\gamma^*$ (see Definition \ref{optimal exponent}).
    Then, let $\widehat{\theta}_n$ be the empirical risk minimizer of the logistic loss:
    \begin{align*}
        \widehat{\theta}_n \defeq \argmin_{\theta \in \mathcal{NN}_{n}} \frac{1}{n}\sum_{i=1}^n l(R_{\varrho,\varrho}(\theta)(X_i),Y_i).
    \end{align*}
    Then the plug-in classification rule based on $\{R_{\varrho, \Tilde{\varrho}}(\widehat{\theta}_n)\}_{n\in\mathbb{N}}$ achieves minimax optimal (up to logarithmic factor) rate of convergence for the excess classification risk. 
\end{theorem}
\begin{remark}
    While condition \eqref{Theorem 5.7: condition 1} is slightly stronger than condition \eqref{Theorem5.5: condition}, we will see from the examples of Section \ref{Section: examples} that the conclusion of Theorem \ref{Theorem: rate of convergence 2} still holds for many interesting function classes.
\end{remark}

\subsection{Two examples} \label{Section: examples}
We demonstrate two applications of Theorem \ref{Theorem: rate of convergence 1} and Theorem \ref{Theorem: rate of convergence 2} to classical function spaces whose Kolmogorov-Donoho optimal exponents are known and are optimally representable by neural networks. 

\subsubsection{H\"older functions}
For a real number $\beta\geq1$, let $m=\floor{\beta}$. We define H\"older class $C^{\beta}([0,1]) \defeq C^{m,\beta-m}([0,1])$ following the definition in Section \ref{Subsection: notations}. We take $\mathcal{F}$ to be the unit ball of H\"older functions. The Kolomogrov-Donoho optimal exponent is given by $\gamma^* = \beta$ and it is optimally representable by neural networks \cite{elbrachter2021deep}. Under certain regularity conditions (Definition 2.2 of \cite{audibert2007fast}) on the marginal distribution of $X$ that is stronger than Assumption \ref{Assumption: 2}, the minimax optimal rate is given by $m^* = \frac{\beta(1+\alpha)}{2\beta + d}$. Then, it suffices to check assumptions \eqref{Theorem5.5: condition} and \eqref{Theorem 5.7: condition 1}, which translate to
\begin{align*}
    \beta-1\geq \frac{\alpha}{2}(1+2\beta);\\
    \beta-1\geq \alpha(1+\beta)
\end{align*}
respectively.
This shows that for ``difficult" problems ($\alpha<1, \beta>1$), the proposed neural network classification rules from  Theorem \ref{Theorem: rate of convergence 1} and Theorem \ref{Theorem: rate of convergence 2} achieve minimax optimal rate of convergence.

\subsubsection{Besov functions} \label{Section: besov functions}
We take $\mathcal{F}$ to be the unit ball of the Besov class $B^m_{2,q}([0,1]^d) \subset L^2([0,1]^d)$ of Besov functions (see Chapter 4.3 of \cite{gine2021mathematical} for a definition and basic properties). Then, $\gamma^* = \frac{m}{d}$ as shown in Theorem 1.3 of \cite{grohs2023phase}. Under the assumption that the density of marginal distribution of $X$ is upper bounded by a constant larger than $1$, which is clearly implied by Assumption \ref{Assumption: 2}, we have $m^* = \frac{m}{2m+d}$ as long as $\alpha=0$ (making Assumption \ref{Assumption: 1} null), $\frac{m}{d}>\frac{1}{q}-\frac{1}{2}$ and $1\leq q\leq \infty$ (see page 2278 of \cite{yang1999minimax}). Assumption \eqref{Theorem5.5: condition}, \eqref{Theorem 5.7: condition 1} translate to
\begin{align*}
    2(m+d)&\geq 2d;\\
    (m+d)&\geq 2d.
\end{align*}
respectively. Note Assumption \eqref{Theorem5.5: condition} is vacuous in this case.
This implies that the conclusion of Theorem \ref{Theorem: rate of convergence 1} holds for all choices of $\alpha,d,q$ satisfying $\frac{m}{d}>\frac{1}{q}-\frac{1}{2}$ while the conclusion of  Theorem \ref{Theorem: rate of convergence 2} holds under the additional assumption $m\geq d$.

\printbibliography

@article{audibert2007fast,
    author = {Jean-Yves Audibert and Alexandre B. Tsybakov},
    title = {{Fast learning rates for plug-in classifiers}},
    volume = {35},
    journal = {The Annals of Statistics},
    number = {2},
    publisher = {Institute of Mathematical Statistics},
    pages = {608 -- 633},
    year = {2007},
    doi = {10.1214/009053606000001217},
    URL = {https://doi.org/10.1214/009053606000001217}
}

@book{steinwart2008support,
  title={Support vector machines},
  author={Steinwart, Ingo and Christmann, Andreas},
  year={2008},
  publisher={Springer Science \& Business Media}
}

@article{pinkus1999approximation,
  title={Approximation theory of the {MLP} model in neural networks},
  author={Pinkus, Allan},
  journal={Acta numerica},
  volume={8},
  pages={143--195},
  year={1999},
  publisher={Cambridge University Press}
}

@article{sharma2020disjoint,
  title={Disjoint Union Metric and Topological Spaces.},
  author={Sharma, Ram Parkash and Goyal, Nitakshi and others},
  journal={Southeast Asian Bulletin of Mathematics},
  volume={44},
  number={5},
  year={2020}
}

@article{petersen2021topological,
  title={Topological properties of the set of functions generated by neural networks of fixed size},
  author={Petersen, Philipp and Raslan, Mones and Voigtlaender, Felix},
  journal={Foundations of computational mathematics},
  volume={21},
  pages={375--444},
  year={2021},
  publisher={Springer}
}

@article{farago1993strong,
  title={Strong universal consistency of neural network classifiers},
  author={Farag{\'o}, Andr{\'a}s and Lugosi, G{\'a}bor},
  journal={IEEE Transactions on Information Theory},
  volume={39},
  number={4},
  pages={1146--1151},
  year={1993},
  publisher={IEEE}
}

@book{devroye2013probabilistic,
  title={A probabilistic theory of pattern recognition},
  author={Devroye, Luc and Gy{\"o}rfi, L{\'a}szl{\'o} and Lugosi, G{\'a}bor},
  volume={31},
  year={2013},
  publisher={Springer Science \& Business Media}
}

@article{elbrachter2021deep,
  title={Deep neural network approximation theory},
  author={Elbr{\"a}chter, Dennis and Perekrestenko, Dmytro and Grohs, Philipp and B{\"o}lcskei, Helmut},
  journal={IEEE Transactions on Information Theory},
  volume={67},
  number={5},
  pages={2581--2623},
  year={2021},
  publisher={IEEE}
}

@article{yang1999minimax,
  title={Minimax nonparametric classification. {I}. Rates of convergence},
  author={Yang, Yuhong},
  journal={IEEE Transactions on Information Theory},
  volume={45},
  number={7},
  pages={2271--2284},
  year={1999},
  publisher={IEEE}
}

@article{onoyama1984limit,
  title={Limit distribution of the minimum distance between independent and identically distributed d-dimensional random variables},
  author={Onoyama, Takuji and Sibuya, Masaaki and Tanaka, Hiroshi},
  journal={Statistical Extremes and Applications},
  pages={549--562},
  year={1984},
  publisher={Springer}
}

@article{rmi/1236864105,
author = {Charles
 
Fefferman},
title = {{Extension of $C^{m, \omega}$-Smooth Functions by Linear Operators}},
volume = {25},
journal = {Revista Matemática Iberoamericana},
number = {1},
publisher = {Real Sociedad Matemática Española},
pages = {1 -- 48},
year = {2009},
}

@article{yarotsky2020phase,
  title={The phase diagram of approximation rates for deep neural networks},
  author={Yarotsky, Dmitry and Zhevnerchuk, Anton},
  journal={Advances in neural information processing systems},
  volume={33},
  pages={13005--13015},
  year={2020}
}

@book{koltchinskii2011oracle,
  title={Oracle inequalities in empirical risk minimization and sparse recovery problems: {\'E}cole D’{\'E}t{\'e} de Probabilit{\'e}s de Saint-Flour XXXVIII-2008},
  author={Koltchinskii, Vladimir},
  volume={2033},
  year={2011},
  publisher={Springer Science \& Business Media}
}

@article{bartlett2019nearly,
  title={Nearly-tight VC-dimension and pseudodimension bounds for piecewise linear neural networks},
  author={Bartlett, Peter L and Harvey, Nick and Liaw, Christopher and Mehrabian, Abbas},
  journal={The Journal of Machine Learning Research},
  volume={20},
  number={1},
  pages={2285--2301},
  year={2019},
  publisher={JMLR. org}
}

@book{gine2021mathematical,
  title={Mathematical foundations of infinite-dimensional statistical models},
  author={Gin{\'e}, Evarist and Nickl, Richard},
  year={2021},
  publisher={Cambridge university press}
}

@article{grohs2023phase,
  title={Phase transitions in rate distortion theory and deep learning},
  author={Grohs, Philipp and Klotz, Andreas and Voigtlaender, Felix},
  journal={Foundations of Computational Mathematics},
  volume={23},
  number={1},
  pages={329--392},
  year={2023},
  publisher={Springer}
}

@article{donoho1998data,
  title={Data compression and harmonic analysis},
  author={Donoho, David L. and Vetterli, Martin and DeVore, Ronald A. and Daubechies, Ingrid},
  journal={IEEE transactions on information theory},
  volume={44},
  number={6},
  pages={2435--2476},
  year={1998},
  publisher={IEEE}
}

@article{devore2021neural,
  title={Neural network approximation},
  author={DeVore, Ronald and Hanin, Boris and Petrova, Guergana},
  journal={Acta Numerica},
  volume={30},
  pages={327--444},
  year={2021},
  publisher={Cambridge University Press}
}

@article{mammen1999smooth,
  title={Smooth discrimination analysis},
  author={Mammen, Enno and Tsybakov, Alexandre B},
  journal={The Annals of Statistics},
  volume={27},
  number={6},
  pages={1808--1829},
  year={1999},
  publisher={Institute of Mathematical Statistics}
}

@article{tsybakov2004optimal,
  title={Optimal aggregation of classifiers in statistical learning},
  author={Tsybakov, Alexander B},
  journal={The Annals of Statistics},
  volume={32},
  number={1},
  pages={135--166},
  year={2004},
  publisher={Institute of Mathematical Statistics}
}

@article{kerkyacharian2014optimal,
  title={Optimal exponential bounds on the accuracy of classification},
  author={Kerkyacharian, Gerard and Tsybakov, Alexandre B and Temlyakov, Vladimir and Picard, Dominique and Koltchinskii, Vladimir},
  journal={Constructive Approximation},
  volume={39},
  pages={421--444},
  year={2014},
  publisher={Springer}
}

@article{hinrichs2008degree,
  title={On the degree of compactness of embeddings between weighted modulation spaces},
  author={Hinrichs, Aicke and Piotrowska, Iwona and Piotrowski, Mariusz and others},
  journal={Journal of Function Spaces},
  volume={6},
  pages={303--317},
  year={2008},
  publisher={Hindawi}
}

@article{petersen2018optimal,
  title={Optimal approximation of piecewise smooth functions using deep ReLU neural networks},
  author={Petersen, Philipp and Voigtlaender, Felix},
  journal={Neural Networks},
  volume={108},
  pages={296--330},
  year={2018},
  publisher={Elsevier}
}

@article{zhang2004statistical,
  title={Statistical behavior and consistency of classification methods based on convex risk minimization},
  author={Zhang, Tong},
  journal={The Annals of Statistics},
  volume={32},
  number={1},
  pages={56--85},
  year={2004},
  publisher={Institute of Mathematical Statistics}
}

@article{bartlett2006convexity,
  title={Convexity, classification, and risk bounds},
  author={Bartlett, Peter L and Jordan, Michael I and McAuliffe, Jon D},
  journal={Journal of the American Statistical Association},
  volume={101},
  number={473},
  pages={138--156},
  year={2006},
  publisher={Taylor \& Francis}
}

@article{boucheron2005theory,
  title={Theory of classification: A survey of some recent advances},
  author={Boucheron, St{\'e}phane and Bousquet, Olivier and Lugosi, G{\'a}bor},
  journal={ESAIM: probability and statistics},
  volume={9},
  pages={323--375},
  year={2005},
  publisher={EDP Sciences}
}

@article{zhang2023classification,
  title={Classification with Deep Neural Networks and Logistic Loss},
  author={Zhang, Zihan and Shi, Lei and Zhou, Ding-Xuan},
  journal={arXiv preprint arXiv:2307.16792},
  year={2023}
}

@article{kim2021fast,
  title={Fast convergence rates of deep neural networks for classification},
  author={Kim, Yongdai and Ohn, Ilsang and Kim, Dongha},
  journal={Neural Networks},
  volume={138},
  pages={179--197},
  year={2021},
  publisher={Elsevier}
}

@article{barron1993universal,
  title={Universal approximation bounds for superpositions of a sigmoidal function},
  author={Barron, Andrew R},
  journal={IEEE Transactions on Information theory},
  volume={39},
  number={3},
  pages={930--945},
  year={1993},
  publisher={IEEE}
}

@article{kohler2022rate,
  title={On the rate of convergence of image classifiers based on convolutional neural networks},
  author={Kohler, Michael and Krzy{\.z}ak, Adam and Walter, Benjamin},
  journal={Annals of the Institute of Statistical Mathematics},
  volume={74},
  number={6},
  pages={1085--1108},
  year={2022},
  publisher={Springer}
}

@article{kohler2020statistical,
  title={Statistical theory for image classification using deep convolutional neural networks with cross-entropy loss},
  author={Kohler, Michael and Langer, Sophie},
  journal={arXiv preprint arXiv:2011.13602},
  year={2020}
}

@article{bos2022convergence,
  title={Convergence rates of deep ReLU networks for multiclass classification},
  author={Bos, Thijs and Schmidt-Hieber, Johannes},
  journal={Electronic Journal of Statistics},
  volume={16},
  number={1},
  pages={2724--2773},
  year={2022},
  publisher={The Institute of Mathematical Statistics and the Bernoulli Society}
}

@misc{suh2024survey,
      title={A Survey on Statistical Theory of Deep Learning: Approximation, Training Dynamics, and Generative Models}, 
      author={Namjoon Suh and Guang Cheng},
      year={2024},
      eprint={2401.07187},
      archivePrefix={arXiv},
      primaryClass={stat.ML}
}

@article{cao2022benign,
  title={Benign overfitting in two-layer convolutional neural networks},
  author={Cao, Yuan and Chen, Zixiang and Belkin, Misha and Gu, Quanquan},
  journal={Advances in neural information processing systems},
  volume={35},
  pages={25237--25250},
  year={2022}
}

@InProceedings{pmlr-v178-frei22a,
  title = 	 {Benign Overfitting without Linearity: Neural Network Classifiers Trained by Gradient Descent for Noisy Linear Data},
  author =       {Frei, Spencer and Chatterji, Niladri S and Bartlett, Peter},
  booktitle = 	 {Proceedings of Thirty Fifth Conference on Learning Theory},
  pages = 	 {2668--2703},
  year = 	 {2022},
  editor = 	 {Loh, Po-Ling and Raginsky, Maxim},
  volume = 	 {178},
  series = 	 {Proceedings of Machine Learning Research},
  month = 	 {02--05 Jul},
  publisher =    {PMLR},
  url = 	 {https://proceedings.mlr.press/v178/frei22a.html},
}

@inproceedings{kou2023benign,
  title={Benign overfitting in two-layer ReLU convolutional neural networks},
  author={Kou, Yiwen and Chen, Zixiang and Chen, Yuanzhou and Gu, Quanquan},
  booktitle={International Conference on Machine Learning},
  pages={17615--17659},
  year={2023},
  organization={PMLR}
}

@article{radhakrishnan2023wide,
  title={Wide and deep neural networks achieve consistency for classification},
  author={Radhakrishnan, Adityanarayanan and Belkin, Mikhail and Uhler, Caroline},
  journal={Proceedings of the National Academy of Sciences},
  volume={120},
  number={14},
  pages={e2208779120},
  year={2023},
  publisher={National Acad Sciences}
}

\newpage
\appendix

\section[Appendix]{Discussion on the relationship between regression and classification}\label{regression and classification}
Here we review some well-known results on the connection between regression and classification and discuss some subtleties in the minimax regime. In this discussion, the domain of $X$ will be $\mathbb{R}^d$ instead of $[0,1]^d$.

Denote by $E_n$ the expectation with respect to the distribution of $Z_1,\dots, Z_n$ and $\mu_X$ the distribution on $\mathbb{R}^d$ induced by $P$ and $X$. In the following, assume that $\{g_n\}_{n\in\mathbb{N}}$ is a plug-in classification rule based on real-valued function sequence $\{f_n\}_{n\in\mathbb{N}}$. We can appeal to Fubini's theorem since all measures are finite and functions are bounded and deduce the following:
\begin{align*}
    E[L(g_n)] - L^* &= E_n[L(g_n) - L(g^*)]\\
    &= E_n[E[ \mathbbm{1}_{g_n(X,Z_1,\dots,Z_n)\neq Y} - \mathbbm{1}_{g^*(X) \neq Y} | Z_1,\dots, Z_n]]\\
    &= E_n\left[ \int_{\mathbb{R}^d} \eta(x) \left(\mathbbm{1}_{g_n(\cdot,Z_1,\dots,Z_n)=0}(x) - \mathbbm{1}_{g^*(\cdot)=0}(x)\right) \mu_X(dx) \right]\\
    &\quad + E_n\left[ \int_{\mathbb{R}^d} (1-\eta(x)) \left(\mathbbm{1}_{g_n(\cdot,Z_1,\dots,Z_n)=1}(x) - \mathbbm{1}_{g^*(\cdot)=1}(x) \mu_X(dx)\right)  \right]\\ 
    &= E_n \left[ \int_{\mathbb{R}^d} |2\eta(x)-1|\mathbbm{1}_{g_n(\cdot,Z_1,\dots,Z_n)\neq g^*(\cdot)}(x) \mu_X(dx)\right]\\
    &\leq E_n \left[\int_{\mathbb{R}^d}2|\eta(x) - f_n(x,Z_1,\dots,Z_n)| \mu_X(dx) \right]\\
    &\leq 2E_n \left[\sqrt[p]{\int_{\mathbb{R}^d}|\eta(x) - f_n(x,Z_1,\dots,Z_n)|^p \mu_X(dx) }\right]
\end{align*}
for any $p\geq 1$ where the second to last inequality follows from the observation that for $x$ such that $g_n(x,X_1,\dots,X_n)\neq g^{*}(x)$, we must have either $f_n(x,X_1,\dots,X_n) < \frac{1}{2} \leq \eta(x)$ or $\eta(x) < \frac{1}{2} \leq f_n(x,X_1,\dots,X_n)$ so that $|\eta(x)-\frac{1}{2}|\leq |\eta(x)-f_n(x,X_1,\dots,X_n)|$, and the last inequality follows from H\"{o}lder's inequality.
In view of the above inequality, fixing $z_1,\dots,z_n$, we may consider $\widehat{f}_n \defeq $
$f_n(\cdot, z_1,\dots,z_n): \mathbb{R}^d \rightarrow \mathbb{R}$ as an approximating function of true $\eta$ corresponding to some unknown $P$ in the $L^p$ sense, and obtain a convergence rate for the excess risk from that of $E_n[\norm{\eta - f_n(\cdot, Z_1,\dots, Z_n)}_{L^p(\mathbb{R}^d,\mu)}]$ for some integer $p\geq 1$. By abuse of notation, we will write this also as $E_n[\norm{\eta - f_n}_{p}]\defeq E_n[\norm{\eta - f_n}_{L^p(\mathbb{R}^d,\mu)}]$, omitting the dependence of $f_n$ on $Z_1,\dots, Z_n$.

More can be said if $p>1$. For a fixed $P$, if $\norm{\eta-f_n}_p \rightarrow 0$ in probability, we have $\rho_n(P) \defeq \frac{E[L(g_n)]-L^*}{E_n[\norm{\eta-f_n}_p]} \rightarrow 0$ as $n \rightarrow \infty$, which means the excess risk converges to $0$ faster than the $L^p$-risk (Theorem 6.5 of \cite{devroye2013probabilistic}). In this sense, classification is easier than regression. Then, a natural question to ask is what can be said about the convergence rate of this ratio. One answer is that no universal (in both $P$ and estimator sequence) bound is possible on this ratio: precisely, given any sequence of numbers converging to $0$ arbitrarily slowly, one can construct some $P$, and a rule $g_n$ based on $f_n$ such that $\norm{\eta-f_n}_p \rightarrow 0$ in probability holds, but the ratio $\frac{E[L(g_n)]-L^*}{E_n[\norm{\eta-f_n}_p]}$ approaches $0$ as slow as the given sequence (see Chapter 6 of \cite{devroye2013probabilistic}). 

On the other hand, if one assumes that either $\eta$ is bounded away from $\frac{1}{2}$ or $L^*=0$, which is a favorable situation for classification, the excess risk can be shown to converge to $0$ at least as fast the $p$th power of the $L^p$-risk (which is smaller than the $L^p$-risk):  $$\frac{E[L(g_n)]-L^*}{E_n[\norm{\eta-f_n}_p^p]} = O(1).$$
Under a less stringent condition than requiring $\eta$ be bounded away from $1/2$, known as the Tsybakov noise condition parametrized by $C_0, \alpha$ (see \eqref{noise condition}), we have for $1\leq p < \infty$,
\begin{align}\label{classification and Lp norm}
    \frac{E[L(g_n)]-L^*}{E_n\left[\norm{\eta-f_n}_p^{\frac{p(1+\alpha)}{p+\alpha}}\right]} \leq C
\end{align}
where $C$ only depends on $C_0,\alpha, p$. See Lemma 5.2 of\cite{audibert2007fast} for a proof.

The discussion in the previous paragraph allows one to derive uniform convergence rates from the approximation properties of $f_n$ for $\eta$. 
While it is well-known that no universal convergence rates are possible, if we restrict $\eta$ to belong to some known family of functions that can be uniformly approximated by a certain class of functions, uniform convergence rates are attainable. 
This is the view we worked with when deriving convergence rate results in Section \ref{Section: convergence rates}.

There is one sense in which the convergence rate of $E[L(g_n)]-L^*$ matches that of $(E_n[\norm{\eta-f_n}_p^p])^{\frac{1}{p}}$.
It is shown in \cite{yang1999minimax} that the minimax rates of $L^2$ risk for a certain class of distributions (characterized by nonparametric classes of functions and regularity conditions on the marginal distribution of $X$) decay to $0$ at the same rate as the minimax rate of the excess risk. Precisely, for some class of probability measures, denoted $\mathcal{P}$,
\begin{align}\label{yang minimax}
    \inf_{f_n}\sup_{P \in \mathcal{P}} (E_n[\norm{f_n-\eta}_2^2])^{\frac{1}{2}} \approx \inf_{g_n}\sup_{P \in \mathcal{P}} E[L(g_n)] - L^*.
\end{align}
where the infimum on the left-hand side is taken over all measurable real-valued functions and the infimum on the right-hand side is taken over all valid plug-in classifiers. 

It is important to observe a key difference from the discussion of the preceding paragraph where we compared the $L^2$-risk associated with a real-valued function $f$ with the classification risk of the plug-in rule associated with the same $f$ (pointwise comparison): in contrast, the classifier that achieves (or nearly so) the infimum of the right-hand side of \eqref{yang minimax} is not necessarily that formed as a plug-in rule of the function that achieves (or nearly so) the infimum of the left-hand side of \eqref{yang minimax}. 

The lesson is that in this uniform regime of minimax risk, we observe a different asymptotic connection between classification and regression than in the pointwise regime: while in the pointwise regime, $E[L(g_n)] - L^*$ converges at least as fast as $E_n[\norm{f_n-\eta}_2^2]$, which implies faster rate than $(E_n[\norm{f_n-\eta}_2^2])^{\frac{1}{2}}$ since $f_n,\eta$ can be assumed to be bounded by 1, in the minimax sense, $E[L(g_n)] - L^*$ converges at the same speed as $(E_n[\norm{f_n-\eta}_2^2])^{\frac{1}{2}}$. 

\section{Proofs}
\subsection{Proof of Lemma \ref{measurability lemma}}\label{proof1}
\begin{proof}
    Under the axiom of countable choice, $B$ is second-countable. Furthermore, it is normal as it is metrizable. Then, we use the fact that a regular, second-countable space can be embedded as a subspace of $\mathbb{R}^{\mathbb{N}}$ with the product topology. The image of this embedding is compact since $B$ is. 
    From now on, we make this identification up to homeomorphism.
    
    Let $\{f_1,f_2,\dots\}$ be a dense set in $B$ and fix $a\in A$. Define $\Tilde{m}: B\rightarrow \mathbb{R}$ as $\Tilde{m}(f) \defeq \inf\{m(a,f)-m(a,f_n), n \in \mathbb{N}\}$, which is upper-semicontinuous. Then, any $\Tilde{f}$ satisfies $m(a,\Tilde{f}) = \sup_{f\in B}m(a,f)$ if and only if $\Tilde{m}(\Tilde{f})=0$. This shows that for each fixed $a$, the set of maximizers of $m(a,\cdot)$ is given by $B_0 \defeq \Tilde{m}^{-1}(0) = \Tilde{m}^{-1}([0,\infty))$, which is closed and hence compact in $\mathbb{R}^{\mathbb{N}}$. Now let $\pi_n: \mathbb{R}^{\mathbb{N}} \rightarrow \mathbb{R}$ be the projection onto the $n$th coordinate. Then, $\pi_1(B_0)$ is compact in $\mathbb{R}$ so it has a maximum element, say $v_1$. Let $B_1\defeq \pi_1^{-1}(v_1) \cap B_0$, which is clearly non-empty and compact. Then proceed inductively, so that we obtain we obtain a sequence of decreasing sets $B_1 \supset B_2 \supset \dots$ Then, the set $\cap_{n=1}^{\infty} B_n$ is non-empty since each finite intersection is non-empty. Now, if any two elements are in this set, by construction they agree on all the coordinates so they are equal. This shows there is a maximum element $\widehat{f}(a) \in B_0$ in the dictionary order over $\mathbb{R}^{\mathbb{N}}$. Thus, for each $a\in A$, we can assign such $\widehat{f}(a)$ to obtain a well-defined mapping from $A$ to $B$. It only remains to show this map is measurable.

    It suffices to show that each $a \mapsto \pi_i(\widehat{f}(a))$ is measurable for all $i\in \mathbb{N}$. Fix a closed interval $[u,v] \subset \mathbb{R}$ for this. We can consider the function $g_{[u,v]}: A \rightarrow \mathbb{R}$ defined by $g_{[u,v]}(a) = \sup \{m(a, f_n): n\in \mathbb{N} \} - \sup \{m(a, f_n): \pi_i(f_n) \in [u,v], n\in \mathbb{N}\}$. This function is Borel measurable as both infimums are taken only over countably many measurable functions. Then, from the observation that $(\pi_i \circ \widehat{f})^{-1}([u,v]) = g_{[u,v]}^{-1}(0)$ we can conclude that indeed $\pi_i \circ \widehat{f}$ is Borel measurable. 
\end{proof}

\subsection{Proof of Theorem \ref{Theorem: universal consistency}}\label{proof2}
\begin{proof}
    First, we check there are no existence and measurability issues in \eqref{theorem1:eq1}. Suppose some $\pi, c_d$ are given (for now). If we regard $z^n =\{X_i, Y_i\}_{i=1,\dots,n}$ as fixed numbers, clearly there is some $\theta \in \mathcal{NN}^{\pi, S_n}_{d,1}(c_d n)$ achieving the minimum in \eqref{theorem1:eq1} by continuity of the associated maps and compactness of $\mathcal{NN}^{\pi,S_n}_{d,1}(c_d n)$. Denote any choice of such $\theta$ for $z^n$ as $\theta_{z^n}$. Moreover, for $\mathcal{Z}= [0,1]^d\times\{0,1\}$, Lemma \ref{measurability lemma} gives the existence of a Borel-measurable function $\widehat{\theta}_n: (\mathcal{Z}^n, \mathcal{B}(\mathcal{Z}^n)) \rightarrow \mathcal{NN}^{\pi,S_n}_{d,1}(c_d n)$ such that $\widehat{\theta}_n(z^n) = \theta_{z^n}$.
    
    We now claim that there are some $\pi,c_d$ such that $P_nM_{\widehat{\theta}_n} = 0$ so that $P_nM_{\widehat{\theta}_n} \geq P_nM_{\theta}$ for all $\theta \in \mathcal{NN}^{\pi,S_n}_{d,1}(c_d n)$. In other words, for each $n$, the empirical risk minimizer of the logistic loss achieves perfect classification accuracy for the $n$ points. This follows from the fact that there exists some $\pi$, which may be assumed to be increasing, such that there is a realization of some $\Tilde{\theta} \in \mathcal{NN}^{\pi,S_n}_{d,1}(c_d n)$ such that
    \begin{align}\label{Theorem 4.2: eq1}
        l(R_{\varrho}(\Tilde{\theta})(X_i),Y_i)\leq \frac{\log 2}{n}, \quad i=1,\dots,n.
    \end{align}
    Such $\widetilde{\theta}$ can be taken to be either a 1 hidden-layer ReLU neural network with width $n$ (see Theorem 5.1 of \cite{pinkus1999approximation}) or a ReLU neural network with width $3$ and $n-1$ hidden-layers (see Proposition 3.10 of \cite{devore2021neural}). This observation and the definition of $\widehat{\theta}_n$ implies the claim $P_nM_{\widehat{\theta}_n} = 0$.   \\
    Fix any $\epsilon>0$. Let 
    \begin{align*}
        \mathcal{F}_0 \defeq \{f : [0,1]^d \rightarrow \mathbb{R}: &f \text{ is measurable and } f(X) \text{ is a}\\
    &\text{version of } E[Y|X]\}.
    \end{align*}
    Let $M^* \defeq -L^*$, be the negative of the Bayes optimal classification risk. Choose any $f_0 \in \mathcal{F}_0$. We may assume $\norm{f_0}_u\leq 1$. By Lusin's theorem, there exists a continuous function $\widetilde{f}_0$ and a measurable set $E$ with $P(E)<\frac{\epsilon}{2}$ such that on $E^c$, $\widetilde{f}_0 = f_0$ and $\norm{\widetilde{f}_0}_u \leq \norm{f_0}_u$. This guarantees that $$\widetilde{\mathcal{F}}_0 \defeq \{\widetilde{f} \in U(C([0,1]^d)): \exists f\in\mathcal{F}_0 \text{ such that outside a set of measure less than } \frac{\epsilon}{2}, f = \widetilde{f} \}$$
    is non-empty, and the classification risk associated with functions in this class differs from $L^*$ by at most $\frac{\epsilon}{2}$.
    Fix any $\widetilde{f}_0 \in \widetilde{\mathcal{F}}_0$. Define the set $A \defeq \{\theta \in \Theta: M^* - PM_{\theta} \geq \epsilon\}$. Because the mapping $R_{\varrho}:\Theta\rightarrow U(C([0,1]^d))$ is surjective (by for e.g., Theorem 3.1 of \cite{pinkus1999approximation}), there exists $\theta_0 \in \Theta$ such that $R_{\varrho}(\theta_0) = \widetilde{f}_0$ and so $PM_{\theta_0}>PM_{\theta'}$ for all $\theta'\in A$. Then, 
    \begin{align}\label{eq9}
      \limsup_{n\rightarrow\infty}\{\widehat{\theta}_n \in A\} \subseteq \left\{\limsup_{n\rightarrow \infty}\sup_{\theta \in A} P_n M_{\theta} \geq PM_{\theta_0}\right\}.
    \end{align}
    Note the $\limsup$ on the left-hand side of \eqref{eq9} is for a sequence of sets while the $\limsup$ on the right-hand side is for a sequence of real numbers. \eqref{eq9} follows from the fact that $\widehat{\theta}_n \in A$ infinitely often implies $\sup_{\theta \in A}P_nM_{\theta} \geq P_nM_{\widehat{\theta}_n} \geq P_n M_{\theta_0}$ infinitely often. But, $P_n M_{\theta_0} \rightarrow PM_{\theta_0}$ almost surely by the strong law of large numbers.

    Before moving further, we show that the map $\theta \rightarrow PM_{\theta}$ is upper-semicontinuous. We use the following convention for the sign function, which is upper-semicontinuous:
    \begin{align*}
        sgn(x) =
        \begin{cases}
            -1, &\text{ if } x<0;\\
            1, &\text{ if } x\geq0.
        \end{cases}
    \end{align*}
    
    The map defined by $t \mapsto -\mathbbm{1}_{(\infty,0)}(t)$ is also upper-semicontinuous. Let $\mathcal{Z}\defeq [0,1]^d \times\{0,1\}$.
    Then, the mapping defined by the following sequence of compositions is seen to be upper-semicontinuous:
    \begin{align*}
        &M: \mathcal{Z}\times \Theta \rightarrow \{-1,0\},\\
        &(z,\theta) \mapsto (z, R_{\varrho}(\theta)) \mapsto (R_{\rho}(\theta)(x),y) \mapsto (sgn(R_{\varrho}(\theta)(x)),y)\\
        &\mapsto sgn(R_{\varrho}(\theta)(x))(2y-1) \mapsto -\mathbbm{1}_{(-\infty,0)}(sgn(R_{\rho}(\theta)(x))(2y-1)).
    \end{align*}
    The claimed upper-semicontinuity follows from the fact that the composition $f\circ g$ is upper-semicontinuous if either $f$ is upper-semicontinuous and $g$ is continuous or both $f,g$ are upper-semicontinuous with $f$ non-decreasing. In what follows, we will use the notation $M_{\theta}(z) \defeq M(z,\theta)$.
    
    Denote $\Theta_0 \defeq \{\theta \in \Theta: P(M_{\theta})= \sup_{\theta' \in \Theta}P(M_{\theta'})\}$. Here $P(M_{\theta})$ denotes the integral of $M_{\theta}$ as a function of $z$ when $z$ is distributed according to $P$, and from the construction of $M_{\theta}$, it follows that $P(M_{\theta}) = -P(sgn(R_{\varrho}(\theta)(X))\neq 2Y-1)$. 
    Note this is the negative of the classification risk associated with the plug-in classifier based on $R_{\varrho}(\theta)+ 1/2$. We also note this set is non-empty because $\Theta$ is compact and the map $\theta \rightarrow PM_{\theta}$ is upper-semicontinuous, essentially by Fatou's lemma.

    Now returning to the proof, Denote by $M_{U}(z) \defeq \sup_{\theta \in U}M_{\theta}(z)$ for any set $U \subseteq \Theta$. In our case, $M_U(\cdot)$ is also measurable because $R_{\varrho}(U)$ is contained in $U(C(\Omega))$, which is separable. For each $\theta \in A$, there exists some small enough open neighborhood $U^{\theta}$ of $\theta$, such that $PM_{U^\theta} < PM_{\theta_0}$ by upper-semicontinuity of the map $\theta \rightarrow PM_{\theta}$ (checked at the beginning of Section \ref{Section: universal consistency}) and the definition of $\theta_0$ and $A$. Consider the open cover of $A$ by the open sets $\{U^{\theta}: \theta \in A\}$ with the aforementioned property. Since $A$ is a compact subset of $\Theta$, we have a finite subcover, which we denote by $\{U^{\theta_1},\dots, U^{\theta_m}\}$ for some $\theta_1,\dots, \theta_m \in A$, $m\in \mathbb{N}$. With this construction,
    \begin{align*}
        \sup_{\theta \in A}P_nM_{\theta} \leq \max_{i: 1\leq i\leq m}P_nM_{U^{\theta_i}} \xrightarrow[a.s.]{n \rightarrow \infty} \max_{i: 1\leq i\leq m} PM_{U^{\theta_i}} < PM_{\theta_0}.
    \end{align*}
    from which we conclude 
    \begin{align}\label{theorem1:eq 9}
        P\left(\limsup_{n\rightarrow \infty}\sup_{\theta \in A}P_nM_{\theta} < PM_{\theta_0} \right) = 1
    \end{align}
    Thus, the right-hand side of \eqref{eq9} has probability $0$ because of \eqref{theorem1:eq 9}, which implies $\widehat{\theta}_n \in A^c$ eventually with probability 1. Since $\epsilon$ was arbitrary, we conclude that 
    \begin{align*}
        \lim_{n\rightarrow \infty} L(R_{\varrho}(\widehat{\theta}_n)+1/2) \rightarrow L^* \text{ with probability } 1.
    \end{align*}
\end{proof}

\subsection{Proof of Theorem \ref{Theorem: rate of convergence 1}}\label{proof3}
\begin{proof}
    In the definition of $\mathcal{NN}_n$, we may assume without loss of generality that all architectures have bounded widths, which ensures that $\mathcal{NN}_n$ may be viewed as a compact, completely metrizable space. A similar argument as in the proof of Theorem \ref{Theorem: universal consistency} shows that $\widehat{\theta}_n$ is well-defined as a measurable mapping from $\mathcal{Z}^n \rightarrow \mathcal{NN}_n$. 
    Take the standard estimation and approximation error decomposition:
    \begin{align*}
         \underbrace{E[L(R_{\varrho}(\widehat{\theta}_n))] - \inf_{f \in R_{\varrho}(\mathcal{NN}_n)}E[L(f)]}_{\circled{1}} + \underbrace{\inf_{f \in R_{\varrho}(\mathcal{NN}_n)}E[L(f)] - L^*}_{\circled{2}}.
    \end{align*}
    Because $\mathcal{F}$ is optimally representable by neural networks, we have
    \begin{align*}
        \sup_{f \in \mathcal{F}} \inf_{\Phi \in \mathcal{NN}_{n}} \norm{f-R_{\varrho}(\Phi)}_{L^2([0,1]])} \leq C_d n^{-\frac{(2+\alpha)m^*}{2(1+\alpha)\gamma^*}}.
    \end{align*}
    Comparison inequality \eqref{classification and Lp norm} and Assumption \ref{Assumption: 2} then implies that 
    \begin{align*}
        \inf_{\Phi \in \mathcal{NN}_n}E[L(R_{\varrho}(\Phi))] - L^* \leq C_{\alpha,d} n^{-m^*}
    \end{align*}
    where $C_{\alpha,d}$ only depends on $\alpha,d$. This bounds $\circled{2}$.
    For $\circled{1}$, we directly appeal to Theorem 5.8 of \cite{koltchinskii2011oracle} using the fact that the VC-dimension of $R_{\varrho}(\mathcal{NN}_n)$ is bounded by $C_d n^{\frac{(2+\alpha)m^*}{2(1+\alpha)\gamma^*}}\log^p(n+1)$ where $p$ is the degree of $\pi$ (\cite{bartlett2019nearly}). Then assumption \eqref{Theorem5.5: condition} ensures that $\circled{1} \leq C_d n^{-m^*}\log^{p}(n+1)$ where $p$ is the degree of $\pi$. Therefore, we conclude that the plug-in classification rule corresponding to $\{R_{\varrho}(\widehat{\theta}_n)\}_{n\in\mathbb{N}}$ achieves minimax optimal rate of convergence for $\mathcal{P}_{\mathcal{F}}$ up to a polylogarithmic factor.
\end{proof}

\subsection{Proof of Theorem \ref{Theorem: rate of convergence 2}}\label{proof4}
\begin{proof}
    Define $T_n = \min_{\substack{1 \leq i<j\leq n}}|X_i - X_j|$. We first note the useful fact (Theorem 1 of \cite{onoyama1984limit}) that for any fixed measure $P$ and arbitrary $\epsilon>0$, there exists a constant $A>0$ small enough so that $T_n \geq An^{-\frac{2}{d}}$ with probability at least $1-\epsilon$ uniformly in $n$. This allows us to assume without loss of generality that there exists some $A>0$ such that $T_n \geq A n^{-\frac{2}{d}}$ with probability $1$. Indeed, there exists large enough $A_0$ such that for all $n\geq N_0$ for some $N_0\in \mathbb{N}$, $P(T_n < A_0n^{-\frac{2}{d}}) \leq C_d n^{-m^*}$. By making $A_0$ large enough, say to $A$, we may assume this holds for all $n\in \mathbb{N}$. 
    
    Fix $n$. Then, for $E\defeq \{X_1,\dots,X_n\}$, define a function $h:E\rightarrow \mathbb{R}$ by
    \begin{align*}
        h(X_i) \defeq 
        \begin{cases}
            -\log(2^{\frac{1}{n}}-1), &\text{ if } Y_i=1;\\
            \log(2^{\frac{1}{n}}-1), &\text{ if } Y_i=0.
        \end{cases}
    \end{align*}
    Then the empirical error of the logistic loss associated with this function satisfies
    \begin{align*}
        \frac{1}{n} \sum_{i=1}^n \log(1+\exp(h(X_i)(2Y_i-1))) \leq \frac{\log 2}{n}.
    \end{align*}
    The above property ensures that the sign of $h(X_i)$ agrees with the sign of $(2Y_i-1)$ for all $i=1,\dots,n$.
    Hereon, we use the symbol $C_d$ to denote a constant that only depends on the input dimension $d$, but the precise values may vary.
    An application of Feffferman's version of Whitney extension theorem page 5 of \cite{rmi/1236864105} ensures that there is an $C^{1,1}$-extension $H:[0,1]^d \rightarrow \mathbb{R}$ of $h$ whose $C^{1,1}$-norm satisfies $\norm{H}_{C^{1,1}([0,1]^d)} \leq C_dAn^{\frac{2}{d}}\log(n+1)$. We check the two conditions in that theorem: first, fix a point $X_0 \in [0,1]^d$ distinct from $X_1,\dots,X_n$ where $c\defeq \min_{i: 1\leq i \leq n}\norm{X_0 - X_i}>0$ and consider the degree $1$ polynomials 
    \begin{align*}
        P_i(x) = \frac{C_d}{An^{\frac{2}{d}}\log(n+1)}(2Y_i-1)(-\log(2^{\frac{1}{n}}-1))\frac{x-X_0}{X_i-X_0}, \quad i=1,\dots,n,
    \end{align*}
    and note that the magnitude of $P_i$'s as well as their order $1$ partial derivatives are uniformly bounded by $1$ if $C_d$ is appropriately redefined because we have $-\log(2^{\frac{1}{n}}-1) \approx \log(n+1)$ (this checks condition (a)). To check condition (b), note that we may take $\omega(t) = \min\left\{\frac{t}{An^{\frac{2}{d}}}, 1\right\}$ for $t\in[0,1]$ and observe
    \begin{align*}
        \norm{(P_i-P_j)(X_j)}_2 \leq \omega(\norm{X_i-X_j}_2)\norm{X_i-X_j}_2, \quad i,j\in \{1,\dots, n \}
    \end{align*}
    holds by construction. Therefore, there exists a $C^{1,1}([0,1]^d)$ function $\widetilde{H}$ such that 
    \begin{align*}
        \widetilde{H}(X_i) = \frac{C_d}{An^{\frac{2}{d}}\log(n+1)}(2Y_i-1)(-\log(2^{\frac{1}{n}}-1))
    \end{align*}
    whose $C^{1,1}$-norm is bounded by a constant only depending on $d$. Thus, by defining
    \begin{align}\label{theorem1: eq1}
        H \defeq C_dAn^{\frac{2}{d}}\log(n+1) \widetilde{H},
    \end{align}
    we get the desired $C^{1,1}$-extension of $h$.
    Furthermore, a recent result of \cite{yarotsky2020phase} guarantees that for any function $f\in C^{1,1}([0,1]^d)$ whose $C^{1,1}$ norm is bounded by 1, there exists a neural network $\phi$ in $\mathcal{NN}_n$ at most $W$ weights such that 
    \begin{align*}
        \norm{f-R_{\varrho,\widetilde{\varrho}}(\phi)}_{L^{\infty}[0,1]^d} \leq \exp(-C_d W^{-1/2}).
    \end{align*}
    Applying this approximation result with appropriately scaled (by a constant depending only on $d$) $\widetilde{H}$ in place of $f$, we conclude using assumption \eqref{Theorem 5.7: condition 1} that there exists a realization of neural network $\widetilde{\phi}$ in $\mathcal{NN}_n$ such that
    \begin{align}\label{theorem1: eq3}
        \norm{\widetilde{H} - R_{\varrho,\widetilde{\varrho}}(\widetilde{\phi})}_{L^{\infty}[0,1]^d} \leq C_d n^{-\frac{2}{d}}.
    \end{align}
    Inequalities \eqref{theorem1: eq1}, \eqref{theorem1: eq3} imply that 
    \begin{align*}
        \norm{H - R_{\varrho,\widetilde{\varrho}}(\widetilde{\phi})}\leq C_d \log(n+1).
    \end{align*}
    Therefore, with a proper choice of $C_d$ in the definition of $\mathcal{NN}_n$, there exists a realization of a neural network in $\mathcal{NN}_n$ 
    whose signs of values at $X_i$'s match the respective signs of $(2Y_i-1)$'s. Thus, we have shown that there exists a realization of some $\Phi_n \in \mathcal{NN}_n$ whose corresponding plug-in rule achieves $0$ empirical error for the classification loss. This implies that $\widehat{\theta}_n$ is also the empirical risk minimizer for the classification loss. 
    
    The rest of the proof is almost the same as the proof of Theorem \ref{Theorem: rate of convergence 1}. Again, take the standard estimation and approximation error decomposition:
    \begin{align*}
         \underbrace{E[L(R_{\varrho,\widetilde{\varrho}}(\widehat{\theta}_n))] - \inf_{f \in R_{\varrho,\widetilde{\varrho}}(\mathcal{NN}_n)}E[L(f)]}_{\circled{1}} + \underbrace{\inf_{f \in R_{\varrho}(\mathcal{NN}_n)}E[L(f)] - L^*}_{\circled{2}}.
    \end{align*}
    Because $\mathcal{F}$ is optimally representable by neural networks, noting that 
    $\mathcal{NN}^{\pi}_{d,1}\left(C_dn^{\frac{(2+\alpha)m^*}{2(a+\alpha)\gamma^*}}\right) \subset \mathcal{NN}_{n}$, we have
    \begin{align*}
        \sup_{f \in \mathcal{F}} \inf_{\Phi \in \mathcal{NN}_{n}} \norm{f-R_{\varrho,\widetilde{\varrho}}(\Phi)}_{L^2([0,1]^d)} \leq C_d n^{-\frac{(2+\alpha)m^*}{2(a+\alpha)}}.
    \end{align*}
    Comparison inequality from \eqref{classification and Lp norm} and Assumption \ref{Assumption: 2} then implies that 
    \begin{align*}
        \inf_{\Phi \in \mathcal{NN}_n}E[L(R_{\varrho,\widetilde{\varrho}}(\Phi))] - L^* \leq C_{\alpha,d} n^{-m^*}
    \end{align*}
    where $C_{\alpha,d}$ only depends on $\alpha,d$. This bounds $\circled{2}$.
    For $\circled{1}$, we directly appeal to Theorem 5.8 of \cite{koltchinskii2011oracle} using the fact that the VC-dimension of $R_{\varrho,\widetilde{\varrho}}(\mathcal{NN}_n)$ is bounded by $C_d n^{\frac{(2+\alpha)m^*}{(1+\alpha)\gamma^*}}$ (\cite{bartlett2019nearly}). Then assumption \eqref{Theorem 5.7: condition 1} ensures that $\circled{1} \leq C_d n^{-m^*}$. Therefore we conclude that the plug-in classification rule corresponding to $\{R_{\varrho,\widetilde{\varrho}}(\widehat{\theta}_n)\}_{n\in\mathbb{N}}$ achieves minimax optimal rate of convergence for $\mathcal{P}_{\mathcal{F}}$ up to a logarithmic factor.
\end{proof}


\end{document}